# ROMEMES: A MULTIMODAL MEME CORPUS FOR THE ROMANIAN LANGUAGE

VASILE PĂIȘ, SARA NIȚĂ, ALEXANDRU-IULIUS JERPELEA, LUCA PANĂ, ERIC CUREA

*Research Institute for Artificial Intelligence "Mihai Drăgănescu", Romanian Academy, Bucharest, Romania*
vasile@racai.ro

## Abstract

Memes are becoming increasingly more popular in online media, especially in social networks. They usually combine graphical representations (images, drawings, animations or video) with text to convey powerful messages. In order to extract, process and understand the messages, AI applications need to employ multimodal algorithms. In this paper, we introduce a curated dataset of real memes in the Romanian language, with multiple annotation levels. Baseline algorithms were employed to demonstrate the usability of the dataset. Results indicate that further research is needed to improve the processing capabilities of AI tools when faced with Internet memes. [1]

## 1 Introduction

Memes represent visual content obtained by combining one or multiple images with text. In some cases, instead of static images, the meme creators use animations or videos, while in some other cases the image is completely missing, being replaced by a simple background color. Furthermore, sometimes the text is missing, the message being conveyed only by visual elements. Regardless of their form, memes can be found in all types of online media, particularly in social networks. They are usually humorous in nature and are used to convey powerful messages that are spread rapidly by Internet users (Bauckhage, 2011). Memes are usually inspired by events happening at the time of their creation. Thus they can be related to sports events, politics, culture, etc. For example, in a period preceding elections, we expect to see memes related to political parties or candidates. The message can be either positive or negative with regard to the represented subject. Hristova (2014) acknowledges that political imaginaries of the structure and scope of the collective will are intimately tied to iconic images that shore up, reinforce, and make literal and legitimate political visions. During the 2020 US presidential election, the social web was flooded with user-generated Internet memes commenting on one of the most important political events worldwide. Internet memes have become a widespread format through which social media users can express individual opinions (political or otherwise) (Johann, 2022). AI applications trying to analyze the message expressed in memes must employ multiple algorithms, adapted to their multimodal nature. Thus image analysis must be combined with natural language processing. Furthermore, certain memes contain a combination of images, requiring additional segmentation. Results obtained from individual analysis processes must be fused together to obtain the final result.

This paper introduces a new corpus of memes in the Romanian language, called RoMemes[2]. It contains multiple levels of annotations, making it suitable for multiple tasks, such as: text recognition, simple sentiment analysis, complex sentiment analysis, fake image detection, identification of political memes. Our contribution is twofold: first, we create and release the curated corpus of Romanian memes, and second, we perform experiments to demonstrate the usability of the corpus for research purposes and establish baseline results. The rest of the paper is organized as follows: Section 2 presents related work, Section 3 presents the methodology used, Section 4 describes the corpus, Section 5 gives experimental results, and we conclude in Section 6.

---

[2] https://zenodo.org/doi/10.5281/zenodo.13120215

## 2   Related work

Vyalla and Udandarao (2020) acknowledge the increased interest in both meme generation and propagation. They propose a framework for automatically generating memes and introduce a dataset consisting of 1.1 million meme captions from 128 classes. Kiela et al. (2020) focus on hate speech detection in multimodal memes. They acknowledge that seemingly harmless texts and images when combined can become hateful. Thus, unimodal analysis is not enough and multimodal algorithms must be employed. Hwang and Shwartz (2023) introduce the MemeCap dataset, containing 6,384 memes, for the task of meme captioning. For this purpose an AI model is presented with a meme and the associated post title (assuming the meme is extracted from social media) and the model must generate a relevant caption, describing the meaning of the meme. The authors acknowledge the challenging nature of the proposed task, due to the metaphorical nature of the meme content. Tanaka et al. (2022) classified a set of 7,500 memes as being humorous or not. Sharma et al. (2020) describe a SemEval challenge consisting of three subtasks: sentiment (positive, negative, and neutral) analysis of memes, overall emotion (humor, sarcasm, offensive, and motivational) classification of memes, and classifying intensity of meme emotion. The best performances achieved were F1 (macro average) scores of 0.35, 0.51 and 0.32, respectively for each of the three subtasks. For the challenge, 1,000 meme samples were provided as trial data, 6,992 samples as training data while 1,879 samples were used as test data. Sharma et al. (2024) introduced MOOD (Meme emOtiOns Dataset), which embodies six basic emotions, and proposed a neural framework named ALFRED (mechanism emotion-Aware muLtimodal Fusion foR Emotion Detection) for meme emotion detection.

Many datasets, including the one proposed in this paper, suffer from class imbalance. To mitigate this problem, Chawla et al. (2002) proposed SMOTE, a synthetic minority over-sampling technique. This approach uses both the under-sampling of the majority class and a particular kind of over-sampling of the minority class by creating "synthetic" examples. Moreover, frameworks for data augmentation, such as NL-Augmenter (Dhole et al., 2023), have been proposed, allowing automatic enrichment of less represented classes.

Considering the Romanian language, different resources and tools are useful for analyzing the text component of a meme. The Romanian Wordnet (Tufiș et al., 2004) is a large lexical database of nouns, verbs, adjectives, and adverbs, grouped into sets of cognitive synonyms (synsets), each expressing a distinct concept. The Romanian Emotion Lexicon (RoEmoLex v.3) (Lupea and Briciu, 2019) includes associations between a set of words and two sentiment orientations (positivity and negativity) as well as eight basic emotions (anger, anticipation, disgust, fear, joy, sadness, surprise, and trust). It was originally translated from an English version, the NRC Word-Emotion Association Lexicon (EmoLex) (Mohammad and Turney, 2010), but now has extra tags, new words and phrases, and synonyms for the original terms. Ciobotaru and Dinu (2021) created the RED dataset for emotion recognition in Romanian tweets. CoRoLa (Tufiș et al., 2019) is the Reference Corpus for Contemporary Romanian Language. It has been used for creating word embeddings used for text representation (Păiș and Tufiș, 2018). Avram et al. (2022) proposed a distilled BERT (Devlin et al., 2019) model for the Romanian language. Given the short text usually associated with memes, social media datasets and models may be relevant when analyzing memes. In this context, Tăiatu et al. (2023) proposed the RoBERTweet model together with a system that detects emotions more accurately than earlier general-domain Romanian and multilingual language models by utilising the RoBERTweet model. Păiș et al. (2022a, 2022b) introduce the MicroBloggingNERo, a dataset of Romanian short messages, specific to micro-blogging platforms.

# 3 Methodology

The corpus of Romanian memes was created at the Research Institute for Artificial Intelligence "Mihai Drăgănescu" of the Romanian Academy. The activity involved 5 annotators working under the supervision of a senior researcher. The annotators team contained a combination of researchers and young students. Before the start of the activity, an "Annotation guidelines" document was created and shared with the annotators. Furthermore, a series of discussions (online and via email) were used to familiarize the annotators with the task. Throughout the annotation process, communication with the annotators team was maintained through online means. All ambiguities were discussed and the annotation guidelines were updated when needed.

The annotators received access to a personal shared Google Drive folder. Each annotator was responsible for collecting their own memes from social media or through Google searches. They were instructed to select only memes with Romanian language text. Following the gathering of new memes, a Google sheet file was filled with information corresponding to each meme. We gathered the following data:

- *file name* - typically this is the file name generated by the platform hosting the meme. However, in some cases this was renamed by the annotator (for example when the platform always produced files with the same name, such as "image.jpg"). In the case of renaming, the original extension was kept. No conversion took place on the files;
- *source* - the source from where the meme was downloaded. This is not necessarily the first place where the meme was uploaded. As memes tend to become viral, being spread extensively by Internet users, it is difficult to determine the actual source of the meme;
- *URL* - the Internet address associated with the page hosting the meme (the location from where it was downloaded);
- *text* - this is the text from the meme. It was extracted exactly as it appears in the meme, including special characters where these were present. We focused only on the message of the meme, while additional text elements (such as logos, messages appearing on clothing) were not extracted, if they were considered not relevant for the meme;
- *complexity* - this field refers to the complexity of the image. We distinguish between: simple images and combinations of multiple images;
- *real/fake* - indicates the authenticity of the image(s) used for the meme. We distinguish between 3 possible values: Real - the image seems to be a real image repurposed for the meme; Fake - the image is a clear fake (such as elements taken from other images superimposed on a base image or containing drawings); DeepFake - the image is fake, but this is hard to tell without additional world knowledge (it looks like a real image, but the annotator knows it cannot be true);
- *sentiment1* - this field is used for simple sentiment analysis, including 3 classes: Positive - the meme conveys a positive sentiment (generally speaking, or towards a person or group); Negative - the meme conveys a negative sentiment; Neutral: the meme conveys no clear positive or negative sentiment;
- *sentiment2* - this field is used for complex sentiment analysis. We consider the 6 emotion categories described by Parrot (2001): joy, love, fear, anger, surprise, sadness. Memes in general are supposed to be amusing. However, the annotation takes into account the emotion that is conveyed by the actual content (situation, person, animal, etc.) of the meme;
- *political* - identifies political memes (related to political parties or individuals involved in politics).

Each annotator had an initial target of 100 memes. We kept only memes that did not contain specific copyright messages. At the end of the annotation process, 50 memes were randomly selected (10 from each annotator) and distributed to be re-annotated by other team members. This allowed us to compute inter-annotator agreement for the different annotation types. The result is given in Section 4.

## 4 The RoMemes Dataset

Following the collection process, the corpus was processed to eliminate any duplicates or memes containing copyright messages. Furthermore, the files were renamed using a numerical sequence, while keeping the original extension, converted to lowercase. This step was needed because some file names contained very long names or special characters, which may not be usable in all file systems. The dataset comprises three folders and one file: "images" - folder containing image files; "text" - folder containing corresponding text files (each file has the same name as its associated image file, with the .txt extension); "conllup" - folder containing CoNLL-U Plus files with the text tokenized and annotated; "metadata.tsv" - tab separated file, containing metadata and annotations. This file contains the following columns:

- *ID* - the numerical sequence associated with the file (file names are constructed based on this ID and the extension, indicated in the following column for images, or .txt for text). Example: "0030023";
- *Extension* - the original file extension (jpg, jpeg, png, webp);
- *Complexity* - the complexity annotation (the annotations were described in Section 3, above);
- *Real_Fake* - the real/fake annotation;
- *Sentiment1* - simple sentiment annotation;
- *Sentiment2* - complex sentiment annotation, based on 6 emotions;
- *Political* - annotation with "Yes"/"No" values indicating if the meme covers political aspects;
- *Width* - width of the image;
- *Height* - height of the image;
- *Channels* - color channels of the image;
- *Mime* - the mime type associated with the file. This is not always the same as can be deduced from the extension. This indicates a possible problem with the platform hosting the memes, from which the corresponding file was downloaded. For example, 2 files with jpg extension actually have a mime type of "image/png", 1 file with jpg extension has mime type of "image/webp";
- *ImageFileSize* - is the size in bytes of the image file;
- *TextSizeChar* - is the size in characters of the associated text file;
- *TextSizeBytes* - is the size in bytes of the associated text file. The text file is saved in UTF-8 encoding. A difference between the TextSizeChar and TextSizeBytes indicates the presence of multibyte UTF-8 characters (usually meaning the presence of Romanian language diacritic characters, such as "ă", "î");
- *TextUppercase* - contains the value "Yes" if the text contains only uppercase characters or symbols;
- *TextLowercase* - contains the value "Yes" if the text contains only lowercase characters or symbols. If both TextUppercase and TextLowercase contain the value "Yes", then the text is composed only of special characters.

The text part of the corpus was processed in the RELATE platform (Păiș et al., 2020; Păiș, 2020; Păiș et al., 2019) and was automatically segmented, tokenized, and annotated with part of speech tags, lemmas, and dependency parsing information. Basic language processing, within the RELATE platform, was performed using UDPipe (Straka et al., 2016), with a new model (Păiș et al., 2021a) trained on the RoRefTrees (Barbu Mititelu et al., 2018) corpus, version 2.7. This choice is justified by the relative processing speed and accuracy of the results, as indicated in Păiș et al. (2021a). The resulting files were exported in CoNLL-U Plus format[3].

---
[3] https://universaldependencies.org/ext-format.html , accessed: 31st August 2024.

| Indicator | Value | Indicator | Value |
|---|---|---|---|
| Number of memes | 462 | Polarity Negative | 221 |
| Min Width | 259 | Polarity Positive | 89 |
| Max Width | 3,839 | Polarity Neutral | 152 |
| Min Height | 194 | Emotion Anger | 79 |
| Max Height | 3,166 | Emotion Fear | 38 |
| Complexity Simple | 358 | Emotion Joy | 87 |
| Complexity Combination | 104 | Emotion Love | 14 |
| Real Images | 293 | Emotion Sadness | 106 |
| Fake Images | 153 | Emotion Surprise | 138 |
| Deep Fake Images | 16 | Political | 147 |

**Table 1**. Statistics for images and categories of the RoMemes corpus

| Indicator | Value | Indicator | Value |
|---|---|---|---|
| Number of tokens | 5,366 | UPOS Noun | 1,172 |
| Avg. tokens/meme | 11.62 | UPOS PropN | 619 |
| Unique tokens | 2,424 | UPOS Verb | 658 |
| Hapax legomena | 1,831 | UPOS Adj | 157 |
| Sentences | 1,074 | UPOS Adv | 302 |
| Avg. tokens/sentence | 5 | UPOS Pron | 421 |
| Unique lemmas | 1,840 | UPOS Num | 141 |

**Table 2**. Statistics for the text part of the RoMemes corpus

Statistics for the RoMemes dataset are provided in Table 1, for the images and annotations, and in Table 2, for the text part. Images vary in size from very small (194 pixels) to very large (3,839 pixels). For many of the annotation categories, the corpus is balanced. However, a few of the categories have a small number of samples (DeepFake images - 16, Emotion Love - 14). Thus, the corpus may not be suitable for experiments with these particular categories. The meme's text is characterized by very short text, having in average 2.32 sentences/meme, and short sentences (an average of 5 tokens/sentence). The text has a lot of nouns (1,791 considering both UPOS Noun and PropN annotations) and there are "sentences" without verbs (658 verbs compared to 1,074 sentences).

## 5 Experiments

Given the different levels of annotation of the dataset, we performed several experiments in order to set baseline values. These are described in the following subsections, with regard to text extraction, sentiment analysis (including polarity and emotion detection), political memes detection and fake image detection.

## 5.1 Text extraction

Text recognition is an essential first-step when developing AI tools that aim to understand memes, proving to significantly increase performance on comprehension tasks (Priyadarshini and Cotton, 2022). Text in memes tends to be partially structured, usually adhering to certain templates (Schmidt et al., 2020), but the background image on which the caption is placed still poses difficulties for current open Optical Character Recognition (OCR) technologies. To evaluate the accuracy of the OCR outputs, we employ two widely-used metrics: BLEU (Papineni et al., 2002) and ChrF++ (Popovic, 2017) (with the standard word order of 2). Both of these metrics indicate the proportion of similar parts from the automatically and manually extracted meme texts. ChrF++ considers both word and character n-grams, as opposed to BLEU which is calculated as a precision only over n-grams. The OCR engine that we use is Tesseract V5 from Google. Tesseract is a technology based on deep learning models with high accuracy and support for many languages (Zacharias, 2020). We experiment with both tessdata[4] (default) and tessdata-best[5] (highest accuracy, but slowest) LSTM models for the Romanian language. However, initial assumptions are made by Tesseract OCR for text recognition, performing best in certain conditions[6], so preprocessing is needed for our use case. Moreover, the page segmentation mode (PSM) has a significant impact on the quality of the output. After careful finetuning of this parameter, we chose PSM 11 ("Sparse text. Find as much text as possible in no particular order." (Iskandarani, 2015)).

Preprocessing is performed by first upscaling the resolution of all memes which have a resolution lower than the 300 Dots per Inch (DPI) standard quality. Then, the alpha channel is removed. Bilateral filtering is used to reduce noise. Because its weight has an additional component of difference in density (Dias and Lopes, 2023), this filter preserves edges instead of blurring them. We used a neighborhood radius of 5, and a spatial sigma of 2 for our filter. We add a small black 10 by 10 pixels border to account for text at the edge of the image. RGB images are converted to grayscale. We applied binarization with a threshold value of 0.9 on a scale from 0 to 1 (where 0 is equivalent to the black color and 1 to white). Memes are finally deskewed with a threshold of 40%. All the image processing was realized using ImageMagick[7]. Example images, before and after the preprocessing pipeline, are given in Figure 1.

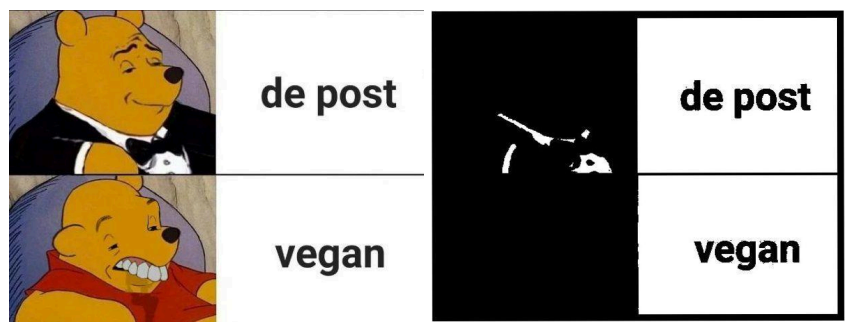

**Figure 1**. A meme about vegan food, before & after our preprocessing

By manually evaluating samples, we observe that most OCR errors come from either: a) lines with random characters in between the lines that actually contain the text of the meme (Table 3), b) misspelling of words caused by failure of OCR to correctly identify some characters. We also note the efficiency of using machine learning models to correct noisy text from OCR (Nguyen et al., 2021), and aim to correct the outputs from our best scoring methodology with two separate experiments: in the first one, we completely remove some errors, and in the second experiment, we aim to modify the text.

---

[4] https://github.com/tesseract-ocr/tessdata , accessed: 31st August 2024.
[5] https://github.com/tesseract-ocr/tessdata_best , accessed: 31st August 2024.
[6] Tesseract documentation. https://tesseract-ocr.github.io/tessdoc/ImproveQuality , accessed: 31st August 2024.
[7] https://imagemagick.org , accessed: 31st August 2024.

We process each line of text, having at most 10 characters (the other lines are kept in order), with a fastText-based language identification model (Joulin et al., 2017). We extract the two languages with the highest probability and remove the line if Romanian is not one of them. The 10 characters limit for the line length was determined through grid search, wanting to achieve the highest BLEU and ChrF++ scores. We tried processing the remaining lines of text through a sequence to sequence transformer model, finetuned for Grammatical Error Correction (GEC) of Romanian text[8]. However, this last step reduced the overall accuracy of the text extraction pipeline. We believe this is due to how certain memes are written: without diacritics, using all caps, with punctuation problems, or grammatically incorrect. Table 3 shows the results from the different text extraction experiments.

| Model | BLEU | ChrF++ |
|---|---|---|
| tessdata + upscale | 26.74 | 63.73 |
| tessdata-best + upscale | 28.52 | 64.60 |
| tessdata + full preprocessing | 37.35 | 65.27 |
| tessdata-best + full preprocessing | 38.06 | 65.43 |
| tessdata-best + preprocess + remove lines | **45.40** | **65.67** |
| tessdata-best + preprocess + remove lines + GEC | 13.78 | 55.23 |

**Table 3**. Text extraction results

**5.2 Sentiment analysis**

We performed sentiment analysis, with regard to both polarity and emotion classification, by employing only the text part of the dataset. We employed the method described in Niță and Păiș (2024). The text was first translated into English using the GoogleTranslator from the deep_translator library to allow compatibility with the NRC emotion lexicon. Given the relatively small dataset, we opted to translate the sentences rather than to translate the lexicon. The text was lemmatised using the WordNet (Fellbaum, 2005) lemmatizer in the NLTK package to help in feature development. Three binary features were created based on the number of words found in the lexicon, indicating a majority of words being positive, negative or neutral in meaning. The same process was utilized in generating the 8 lexicon-predicted emotion features. If an emotion was the most encountered according to the lexicon, it would be recorded as '1' (when a tie occurred, all of the most-encountered emotions were marked as '1'). One aspect to note is that the NRC does not contain data related to how words express 'love'. Thus the value of occurrence for this emotion was interpreted as a mean between 'trust' and 'joy'. This choice was motivated by Robetr Plutchik's general psychoevolutionary theory of emotion (1980), in which, beyond the eight primary emotions: Anger, Anticipation, Disgust, Fear, Joy, Sadness, Surprise, Trust, all other emotions are mixtures, or compounds of these primary emotions, including love, which is a combination of joy and trust. A confidence value was also computed based on the number of words belonging to different classes which were found in the lexicon for a given sample.

Additional features were generated using a BERT model for sentiment and emotion predictions on the translated data, trained by Niță and Păiș (2024). The LLM used for training the system was bert-base-uncased. This was chosen due to the fact that the base and large models produced almost the same results (both F1 scores of around 0.3 for polarity and around 0.16 for emotion classification), and thus the smaller system was chosen. The LLM classifier has two additional linear layers, with 2,048 and 1,024 cells respectively, employing ReLU and tanh activation functions. These are followed by a final class prediction head. The model was trained for at least 5 epochs and a maximum of 20 epochs, with early stopping when there was no improvement for 3 epochs. During the first 3 epochs, the LLM was frozen, and only the last linear layers were trained. A batch size of 6 was used. The learning rates for the LLM and the other layers were kept separate. The best hyperparameters were found to be an encoder learning rate of 1.0e-05 and a linear layers learning rate of 3.0e-05. The final model training lasted 5 epochs.

---

[8] https://huggingface.co/readerbench/RoGEC-mt0-base , accessed: 31st August 2024.

The feature set (dataset features as described in Section 4, lexicon features, and BERT features) was split into training, development, and testing datasets and used in a Random Forest algorithm. To tackle the imbalance of the dataset, new examples were synthesized using a SMOTE pipeline (Chawla et al., 2002), and a Stratified K-fold with n_folds = 5 was used to avoid overfitting. The best hyperparameters were found using GridSearch. Best hyperparameters for the Random Forest used for polarity: max_depth: None, min_samples_leaf: 2, min_samples_split: 2, n_estimators: 200. Best hyperparameters for emotion classification: max_depth: None, min_samples_leaf: 1, min_samples_split: 4, n_estimators: 100. Feature selection was then done, removing some of the features in order to achieve a higher F1 score for the test set, while also making sure that the model is still generalizing well. For both polarity and emotion detection, the BERT predictions for negative polarity and surprise were dropped, along with the feature relating to whether an image is deepfake. Also, the text length was dropped only for emotion classification. For polarity the best results actually occurred when the BERT features were not used. Detailed results are shown in Table 4.

| Model | Polarity | | | | Emotion | | | |
|---|---|---|---|---|---|---|---|---|
| | Acc | P | R | F1 | Acc | P | R | F1 |
| BERT | **0.489** | 0.163 | 0.333 | 0.321 | 0.255 | **0.420** | 0.166 | 0.103 |
| RF without BERT | 0.443 | **0.443** | **0.443** | **0.442** | 0.314 | 0.323 | 0.317 | 0.317 |
| RF with BERT | 0.412 | 0.426 | 0.414 | 0.415 | **0.343** | 0.346 | **0.342** | **0.342** |

Table 4. Sentiment analysis (polarity and emotion detection) results

**5.3 Political memes detection**

Memes have a particular appeal to world-wide cultures, and are being used by political actors to communicate various political messages with the aim of changing certain beliefs (Beskow et al., 2020). We aim to establish a baseline for detecting Romanian language political memes, in which we only make use of the meme's text content, encouraging future multimodal input approaches. Although most Large Language Models (LLMs) are not up to date with the political world stage, depending on their knowledge cutoff, they have proven to be useful in various zero-shot learning classification problems (Chae and Davidson, 2023). We experiment with two instruction LLMs: Llama3.1-8B-Instruct (Dubey et al., 2024) and RoLlama3-8B-Instruct (Masala et al., 2024). We feed the LLM each meme text input and prompt it to classify it as political not. The instruction we used has been carefully tuned and stimulates chain-of-thought reasoning (Wei et al., 2022): *Eşti un asistent util care clasifică conţinutul text al unui 'meme' drept politic sau nu. Îţi explici pe scurt decizia, iar apoi răspunzi la final cu 'DA' sau 'NU'* (english translation: *You are a useful assistant that classifies the textual content of a 'meme' as political or not. You briefly explain your decision, and then answer at the end with 'YES' or 'NO'*). The results are given in Table 5. The low performance seems to suggest the need for future more complex meme comprehension algorithms, employing multimodal data.

| Model | Acc | P | R | F1 |
|---|---|---|---|---|
| Llama3.1-8B-Instruct | 0.52 | 0.35 | 0.57 | 0.43 |
| RoLlama3-8B-Instruct | 0.62 | 0.34 | 0.20 | 0.25 |

Table 5. Political classification based on the memes text (Accuracy, Precision, Recall, F1 scores)

## 5.4 Fake image detection

In this series of experiments, we are concerned with real vs fake classification of the meme image(s). Even though this is an experiment usually conducted only on the image content, we also explore the incorporation of the text content. The intuition behind this approach is that sometimes the text content may contain clues regarding the authenticity of the image. For example, considering the text "Aşa ar arăta dacă…" ("This is what he would look like if…") this may be a clear indication of a fake image. In the first series of experiments we considered a CNN-based Residual network (ResNet50, ResNet101) for image processing. The text was processed through a DistilBERT model (the 'distilbert-base-uncased' variant, using a maximum length of 128 tokens). After extracting features from both the image and text, these features are concatenated. The combined feature vector then passes through a custom classifier. This classifier consists of a linear layer that reduces the combined features to 512 dimensions, followed by a ReLU activation and a dropout layer with a rate of 0.3. A final linear layer then produces the output for binary classification. The training process uses the following parameters (found through grid search): batch size 32, learning rate 0.0005, weight decay 1e-4, training epochs 30. For training we employ the Adam optimizer, binary cross-entropy with logits loss, and a ReduceLROnPlateau learning rate scheduler, with a patience of 3 epochs. Results are given in Table 6, in terms of macro-precision, macro-recall, macro-F1, and accuracy. The best performing model is represented by a ResNet101 image processing model, without text information. Thus it seems that our initial assumption that the text may provide useful information for classification was wrong. Overall it hurts the model's performance and an image-only approach is better.

| Model | P | R | F1 | Acc |
|---|---|---|---|---|
| ResNet50 + text | 0.7198 | 0.7179 | 0.7187 | 0.7246 |
| ResNet101 + text | 0.6215 | 0.6293 | 0.6167 | 0.6232 |
| ResNet50 | 0.7492 | 0.7543 | 0.7503 | 0.7536 |
| ResNet101 | **0.8027** | **0.7727** | **0.7805** | **0.7971** |

Table 6. Results from CNN-based experiments for fake meme image(s) detection

A second series of experiments employed Generative Adversarial Networks (GANs) (Goodfellow et al., 2014). We employ two architectures: CNN-based, using four convolutional layers, with max pooling, batch normalization and ReLU activation, and a pre-trained VGG16 network (Simonyan and Zisserman, 2015), with the weights of the convolutional layers frozen, allowing the rest of the layers to be trained. For training, we employed the Adam optimizer with a learning rate of 0.0001, weight decay of 1e-5, and binary cross-entropy with logits loss. A learning rate scheduler (ReduceLROnPlateau) is used to adjust the learning rate based on the validation loss, with a patience of 5 epochs and a reduction factor of 0.1. The training loop runs for a maximum of 75 epochs, with an early stopping mechanism that has a patience of 10 epochs. Gradient clipping is employed during training, with a maximum norm of 1.0, to prevent exploding gradients and stabilize the training process. This is particularly important given the depth of the network and the potential for unstable gradients in GAN-inspired architectures. For these experiments we did not include text information. Results are given in Table 7, in terms of macro-precision, macro-recall, macro-F1, and accuracy. The VGG16 results are better compared to the basic CNN approach. However, the results are lower than the previous ResNet101 approach.

| Model | P | R | F1 | Acc |
|---|---|---|---|---|
| GAN with CNN | 0.5974 | 0.6000 | 0.5984 | 0.6304 |
| GAN with VGG16 | **0.6693** | **0.6813** | **0.6707** | **0.6848** |

Table 7. Results from GAN experiments for fake meme image(s) detection

# 6   Conclusion and future work

This paper introduced the RoMemes corpus, a curated dataset of Romanian memes, with multiple levels of annotations. The dataset was publicly released. The scripts used for processing the raw corpus and for the baseline experiments are also available inside an associated GitHub repository[9]. Baseline experiments were detailed in Section 5. For all tasks, additional algorithms are needed in order to improve the overall text extraction or classification performance. Future experiments will focus on using the Romanian lexicon RoEmoLex as a replacement for the NRC and eliminating the need for translation when performing sentiment analysis. Furthermore, we'll focus on improving multimodal algorithms for the different classification levels. Of particular interest is the detection of deep fake images used in memes. With the development of generative artificial neural models, computers can generate vivid face images that can easily deceive human beings. These generated fake faces will inevitably bring serious social risks, e.g., fake news, fake evidence, and pose threats to security (Barsh et al., 2020). Our dataset will need to be expanded with additional memes containing deep fake images in order to be successfully employed for the multimodal task of deep fake detection.

---

[9] https://github.com/racai-ai/RoMemes